\documentclass[sigconf,nonacm,natbib=true]{acmart}
\usepackage{multirow}
\usepackage{caption}
\usepackage{balance}
\renewcommand{\paragraph}[1]{\medskip\noindent\textbf{#1:~}}
\usepackage{array}

\newcolumntype{M}[1]{>{\centering\arraybackslash}m{#1}}
\newcolumntype{N}{@{}m{0pt}@{}}
\usepackage[show]{chato-notes}
\usepackage{balance}
\usepackage{siunitx}

\usepackage{tabularx}
\usepackage{makecell}
\usepackage{arydshln}
\usepackage{amsmath}
\usepackage[normalem]{ulem}
\definecolor{aqua}{rgb}{0.0, 1.0, 1.0}
\definecolor{babypink}{rgb}{0.96, 0.76, 0.76}
\definecolor{bananamania}{rgb}{0.98, 0.91, 0.71}
\definecolor{carolinablue}{rgb}{0.6, 0.73, 0.89}
\definecolor{columbiablue}{rgb}{0.61, 0.87, 1.0}

\usepackage{graphicx}
\usepackage{subfig}
\usepackage{xspace}

\newcommand{\ourdataset}{ClariT\xspace} 

\newcommand{\ourmodel}{MAS2S\xspace}
\newcommand{\partitle}[1]{\vspace{2mm}\noindent\textbf{#1}}

\begin{document}

\title{Towards Asking Clarification Questions for Information Seeking on Task-Oriented Dialogues}

\author{Yue Feng}
\affiliation{%
  \institution{University College London}
  \city{London}
  \country{UK}}
\email{yue.feng.20@ucl.ac.uk}

\author{Hossein A. Rahmani}
\affiliation{%
  \institution{University College London}
  \city{London}
  \country{UK}}
\email{hossein.rahmani.22@ucl.ac.uk}

\author{Aldo Lipani}
\affiliation{%
  \institution{University College London}
  \city{London}
  \country{UK}}
\email{aldo.lipani@ucl.ac.uk}

\author{Emine Yilmaz}
\affiliation{%
  \institution{University College London}
  \city{London}
  \country{UK}}
\email{emine.yilmaz@ucl.ac.uk}


\begin{abstract}
Task-oriented dialogue systems aim at providing users with task-specific services.
Users of such systems often do not know all the information about the task they are trying to accomplish, requiring them to seek information about the task.
To provide accurate and personalized task-oriented information seeking results, task-oriented dialogue systems need to address two potential issues: 1) users' inability to describe their complex information needs in their requests; and 2) ambiguous/missing information the system has about the users. 
In this paper, we propose a new Multi-Attention Seq2Seq Network, named MAS2S, which can ask questions to clarify the user's information needs and the user's profile in task-oriented information seeking. 
We also extend an existing dataset for task-oriented information seeking, leading to the \ourdataset which contains about 100k task-oriented information seeking dialogues that are made publicly available\footnote{Dataset and code is available at \href{https://github.com/sweetalyssum/clarit}{https://github.com/sweetalyssum/clarit}.}. Experimental results on \ourdataset show that MAS2S outperforms baselines on both clarification question generation and answer prediction.
\end{abstract}

\maketitle

\section{Introduction}
The primary goal of a task-oriented dialogue system is to help the user complete a task. Since most tasks can be highly complex, and users often lack background knowledge about the tasks, users have to search for task-related information to accomplish tasks. This type of information seeking behaviour is referred to as task-oriented information seeking and is an important problem that needs to be solved by task-oriented dialogue systems~\citep{louvan2020recent,madotto2020learning}.

\begin{figure}[!t]
\centering
\includegraphics[width=0.47\textwidth]{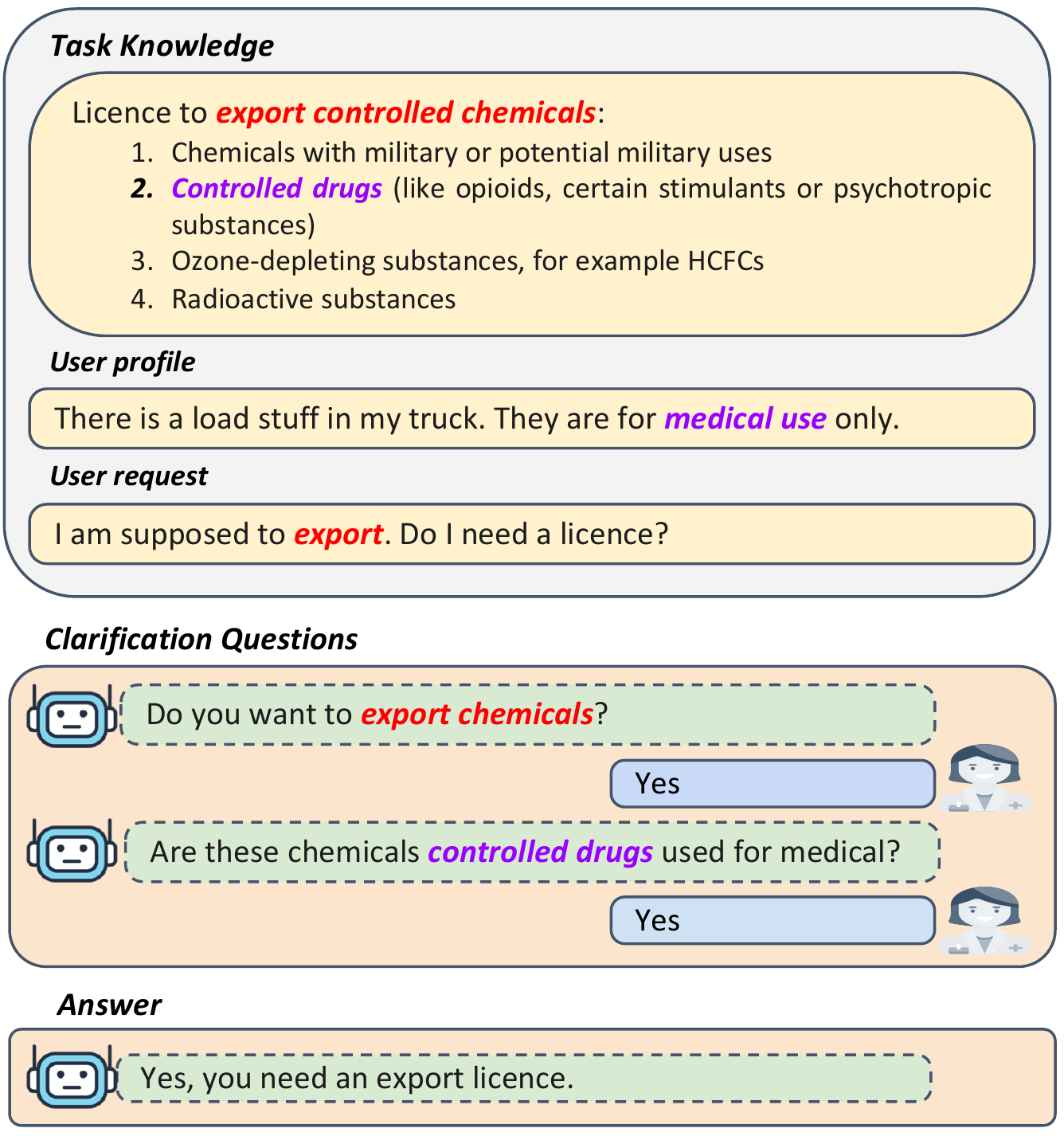}
\caption{An example of task-oriented information seeking. A user needs to search for information related to a specific task. Due to the unclear user request and user profile, the task-oriented dialogue system should ask clarification questions to clarify the user request and the user profile based on the task knowledge to provide the answer.}
\label{fig:example}
\end{figure}

During the information-seeking process in task-oriented dialogues, users often fail to formulate their complex task-related information needs in a single request. 
Furthermore, the system may not have enough information about the profile of the user to accurately respond to the user's request.
In order to provide accurate and personalized task-related information-seeking results, systems have to ask questions to clarify the user request and the user profile. Figure~\ref{fig:example} shows an example of a task-oriented information-seeking. 

\begin{figure*}[!t]
\centering
\includegraphics[width=0.7\textwidth]{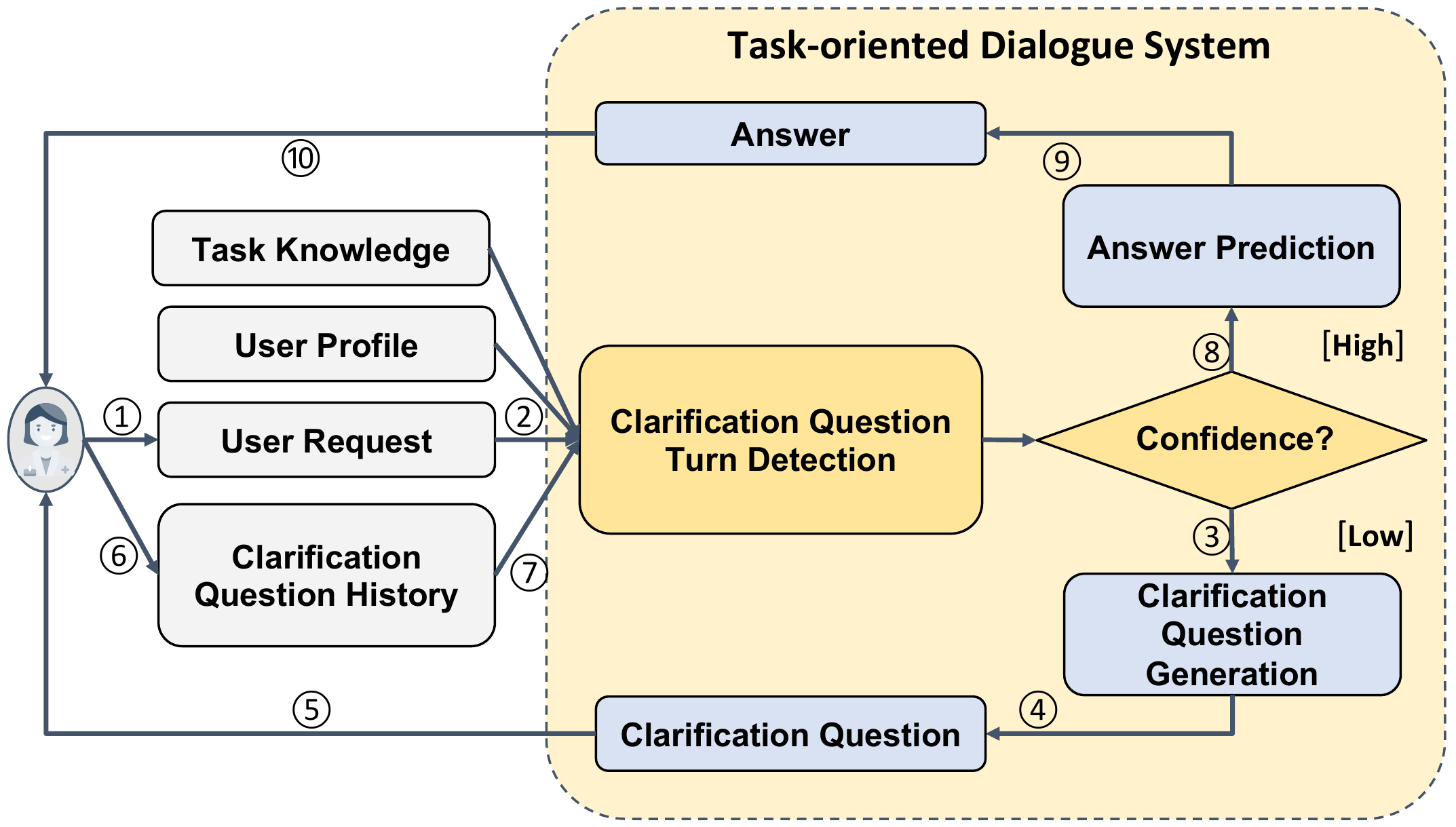}
\caption{The Task-oriented System Ask Paradigm.}
\label{fig:paradigm}
\end{figure*}

Many methods have been proposed for asking clarification questions for general information seeking tasks~\citep{aliannejadi2021building,rao2019answer}, which include generation-based models~\cite{kumar2020clarq,cao2019controlling} and ranking-based models~\cite{xu2019asking,aliannejadi2019asking}. 
Ranking based models assume the existence of a candidate set of clarification questions and cast clarification question generation as a ranking problem. Generation based methods tend to be more challenging as they do not assume the existence of a set of candidate questions.

Compared to asking clarification questions for general information seeking, two issues need to be addressed for task-oriented information seeking: 1) user's information needs are related to a specific task. Existing task-oriented systems assume that the system has access to some prior knowledge about the task~\citep{feng2020doc2dial,madotto2018mem2seq,eric2017key}. Therefore, a task-oriented information-seeking model needs to ask clarification questions considering the task knowledge the system has, and; 
2) for different user information, the answers may be different~\citep{mehrotra2015terms}. Task-oriented dialogue systems should provide personalized answers by incorporating users' profiles. Therefore, a task-oriented information-seeking model also needs to ask clarification questions considering the user profile. 

In this paper, we focus on the problem of clarification question generation for task-oriented information seeking. We propose a \textit{Task-oriented System Ask} paradigm, which considers task knowledge to clarify the user request and user profile. We also propose a Multi-Attention Seq2Seq Network (\ourmodel) as an implementation of this paradigm, which generates clarification questions and answers to the user requests in a single model. Specifically, \ourmodel consists of a text encoder, a final response confidence embedding network, and a natural language decoder.

To the best of our knowledge, no datasets have been developed for task-oriented information seeking, which ask questions to clarify user requests and user profiles based on task knowledge. The closest dataset that can be used for this purpose is the ShARC~\cite{saeidi2018interpretation} dataset, which was built for asking questions to clarify user requests based on task knowledge. However, the ShARC dataset does not contain user profile information, which is critical for task-oriented information seeking. To build a suitable dataset for our setting, we extend dialogues in ShARC dataset with user profiles as a new public dataset, named \ourdataset. 

We compare \ourmodel on \ourdataset with competitive clarification question generation models for general purpose information seeking and natural language generation models in task-oriented dialogue systems. Experiment results show that \ourmodel significantly outperforms all the baselines. Extensive analyses in the Section~\ref{sec:results} also reveal the significance of clarifying user requests and user profiles based on task knowledge in task-oriented information seeking.

\section{Related Work}
\subsection{Task-Oriented Dialogues}
Task-oriented dialogue systems  focus on helping users accomplish different tasks. Traditional systems~\cite{mehrotra2015terms,wen2017network,eric2017key,lei2018sequicity,zhong2019e3,liang2020moss,feng2021sequence} adopt a pipelined approach that requires dialogue state tracking for understanding the user’s goal, dialogue policy learning for deciding which system action to take, and natural language generation for generating system responses. With the emergence of multi-domain task-oriented dialogue datasets~\cite{budzianowski2018multiwoz,shah2018bootstrapping,rastogi2020towards,feng2020doc2dial,gunasekara2020overview}, we are witnessing methods to 
gradually transition from modularized to end-to-end modelling approaches~\cite{budzianowski2019hello,lin2020mintl,hosseini2020simple,peng2021soloist,yang2021ubar}. 

However, the information-seeking problem in task-oriented dialogues is still under-studied, with little research and no datasets available for this problem. As a result, we propose, a framework for task-oriented information seeking, a dataset, and a deep learning solution to solve this task.

\subsection{Clarification Question Generation}
With the emergence of various conversational devices, clarification question generation for information seeking has achieved new attention in recent years. \citet{zhang2018towards} proposed to ask aspect-based clarification questions in the right order so as to understand the user's needs. \citet{aliannejadi2019asking} and \citet{xu2019asking} proposed a ranking model to select clarification questions in open-domain information-seeking conversations. \citet{cao2019controlling} proposed to feed expected question specificity along with the context to generate specific and generic clarifying questions.
\citet{rao2019answer} proposed a sequence-to-sequence generation network using the attention mechanism to generate clarification questions. \citet{kumar2020clarq} generated clarification questions by sampling comments from StackExchange posts. 

To the best of our knowledge, no previous work exists on clarification question generation for information seeking on task-oriented dialogues, which is the focus of this paper.

\begin{figure*}[!t]
\centering
\includegraphics[scale=0.55]{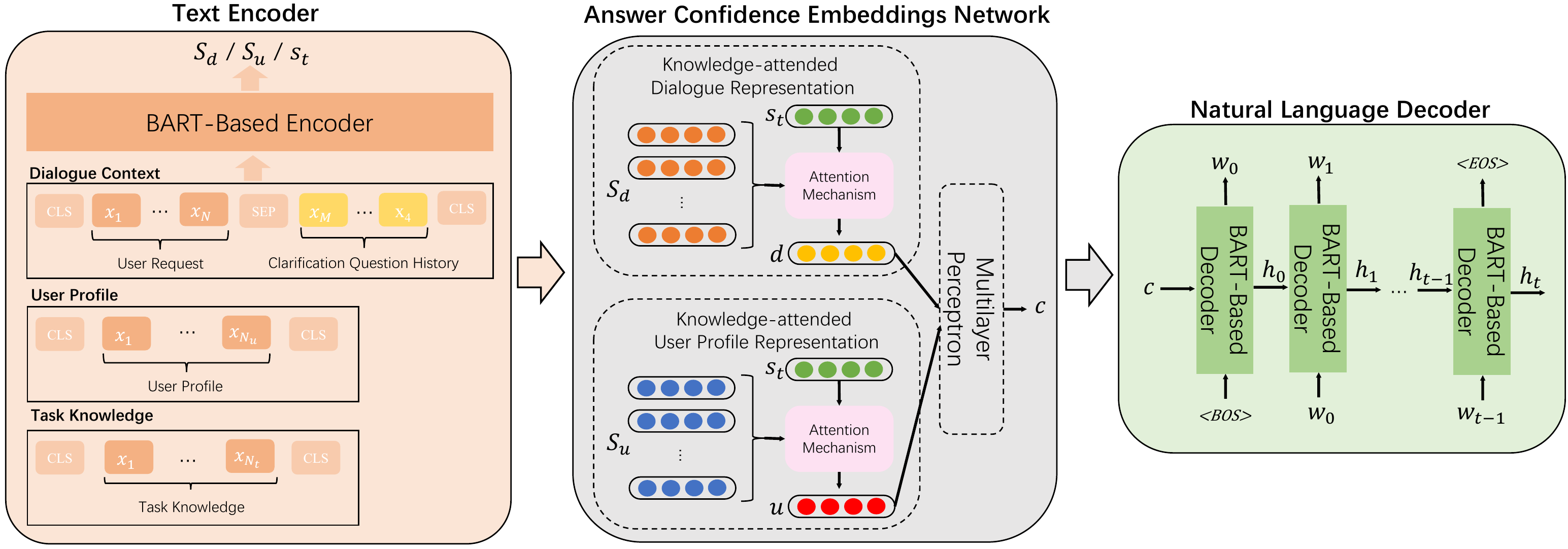}
\caption{The architecture of \ourmodel for information seeking on task-oriented dialogues.}
\label{fig:framework}
\end{figure*}

\section{PROBLEM FORMALIZATION}
\subsection{The Task-oriented System Ask Paradigm}
We propose a clarification question generation paradigm, named Task-oriented System Ask Paradigm, for information seeking in task-oriented dialogues. The workflow of the Task-oriented System Ask Paradigm is depicted in Figure~\ref{fig:paradigm}. After a user provides an initial user request related to a task, the system calculates the answer confidence with the \textit{clarification question turn detection} module based on the user request, the user profile, and the task knowledge. If the system is not sufficiently confident with the answer, it will then generate a clarification question to ask using the \textit{clarification question generation} module, which also considers user request, user profile, and task knowledge. After the user answers the clarification question, the system returns to the previous state, but this time it considers not only the user’s initial request but also the newly collected clarification question and user feedback. This process will continue until the system is confident enough to provide an answer, in which case the system will display the answer to the user. 

\subsection{Notations and Problem Statement}

Figure~\ref{fig:example} shows an example of a task-oriented information-seeking dialogue. A user has an initial user request $R$ that relates to a specific task. This user also has a user profile $U$ that describes the personalized background information. 
If the user request $R$ and user profile $U$ are underspecified, the system cannot provide an accurate and personalized answer $Y$ to the user request $R$. Therefore, 
the system needs to use the task knowledge $T$ to infer a clarification question $Q$ to clarify the user request $R$ and user profile $U$. We thus build the following conversation for this task-oriented information-seeking behaviour,
\begin{gather}
R, U, T | Q_1,A_1,Q_2,A_2, ..., Q_K,A_K | Y,
\end{gather}
where $Q_k$ is a clarification question asked by the system, $A_k$ is the user answer to the question $Q_k$, and $K$ is the number of clarification questions in the dialogue.

Based on the above notation, the task-oriented dialogue system aims at learning models for the following two key tasks:


\partitle{Clarification Question Generation.} 
Given a user request, a user profile, a task knowledge, and a dialogue history, generate the next clarification question to ask. Specifically, a generative model is trained by maximizing the probability of each clarification question in each of the training dialogues:
\begin{multline}
P(Q_{k} | R, U, T, Q_1,A_1, \dots, Q_{k-1},A_{k-1}), \\ k \in \{1,\dots,K\}
\end{multline}

\partitle{Answer Prediction.}
Given a user request, a user profile, a task knowledge, and a dialogue history, generate an answer for the user request. Specifically, a generation model is trained by maximizing the probability of the ground truth answer for each of the training conversations:
\begin{gather}
P(Y | R, U, T, Q_1,A_1, ..., Q_K,A_K)
\end{gather}

\section{Multi-Attention Seq2Seq Networks}
In this section, we propose a Multi-Attention Seq2Seq Network, named MAS2S, that is able to ask clarification questions based on the Task-oriented System Ask paradigm.
\ourmodel formalizes clarification question generation and answer prediction as a sequence-to-sequence problem using BART~\cite{lewis2020bart} and Attention Networks~\cite{vaswani2017attention}. As shown in Figure~\ref{fig:framework}, \ourmodel consists of a text encoder, an answer confidence embedding network, and a natural language decoder. At each turn of the dialogue, the text encoder transforms the user request and the clarification question history into dialogue embeddings using the BART encoder; the text encoder transforms the user profile into the user embeddings using the BART encoder; the text encoder transforms the task knowledge into the knowledge embeddings also using the BART encoder; the answer confidence embeddings network creates knowledge-aware dialogue representations and knowledge-aware user representations using the attention mechanism to calculate answer confidence embeddings; finally, the natural language decoder sequentially generates a clarification question or an answer for the user request on the basis of the answer confidence embeddings. Below we describe the different components of the model in detail.

\subsection{Text Encoder}

The text encoder takes the text of a dialogue, a user profile, and of a task knowledge as input respectively and employs BART to construct the corresponding semantic embeddings. 

More specifically, to generate the semantic embeddings of dialogue context, the BART encoder is given the token sequence $X=(\textsc{[cls]}, x_1, ..., x_N, \textsc{[sep]}, x_1, ..., x_M, \textsc{[cls]})$, which are the sub-word tokens of user request with length $N$ and the clarification question history with length $M$. The \textsc{[cls]} and \textsc{[sep]} are the start-of-text/end-of-text and separator pseudo-tokens. The output embeddings of each token are used as the dialogue semantic embeddings, referred to as $S_d = (d_1, ..., d_{N+M+3} )$.

To generate the semantic embeddings of a user profile, the BART encoder takes a sequence of user profile tokens with length $N_u$ as inputs, denoted as $X=(\textsc{[cls]}, x_1, ..., x_{N_{u}}, \textsc{[cls]})$. The output is a sequence of embeddings with length $N_u+2$, denoted as $S_u = (u_1, ..., u_{N_u+2} )$ and referred to as user profile embeddings, with one embedding for each token.

We also use a BART encoder to generate representations for task knowledge. 
The input is a sequence of task knowledge tokens with length $N_t$, denoted as $X=(\textsc{[cls]}, x_1, ..., x_{N_{t}}, \textsc{[cls]})$. The state of the final \textsc{[cls]} is used as the task knowledge semantic embeddings, referred to as $s_t$. 

\subsection{Answer Confidence Embeddings Network}

The answer confidence embeddings network takes the sequence of dialogue embeddings, the sequence of user profile embeddings, and the task knowledge embeddings as input and first calculates knowledge-attended dialogue representations and knowledge-attended user profile representations. In this way, the semantic information from the dialogue context and user profile is represented based on the task knowledge. Then the answer confidence embeddings can be obtained by the reconstructed knowledge-attended semantic embeddings.

Specifically, we first use the attention mechanism to calculate the knowledge-attended representations between task knowledge $s_t$ and the dialogue $S_d$ / user profile $S_u$ by bilinear interaction: 
\begin{gather}
A_{d} = \text{softmax}(\text{exp}(S_d^\mathsf{T} W_d s_t)), \\
A_{u} = \text{softmax}(\text{exp}(S_u^\mathsf{T} W_u s_t)),
\end{gather}
where $W_d$ and $W_u$ are the bilinear interaction matrices to be learned. Then the knowledge-attended dialogue representations $d$ and the knowledge-attended user profile representations $u$ are calculated as $d = S_d^\mathsf{T} A_{d}$ and $u = S_u^\mathsf{T} A_{u}$.

To obtain the answer confidence embedding $c$ for current dialogues and users, we concatenate the knowledge-attended dialogue representations and the knowledge-attended user profile representations. A multi-layer perceptron derives the answer confidence embedding $c$ by the following equation:
\begin{gather}
c = \text{MLP}([d;u]).
\end{gather}

\subsection{Natural Language Decoder}
The natural language decoder generates clarification questions or answers to the user's request by attending to the answer confidence embeddings. We employ a BART decoder for the natural language decoder, which takes the answer confidence embedding $c$ as its initial hidden state. At each decoding step $t$, the decoder receives the embedding of the previous item $w_{t-1}$, and the previous hidden state $h_{t-1}$, and produces the current hidden state $h_t$:
\begin{equation}
h_t = \text{BART}(w_{t-1}, h_{t-1}).
\end{equation}

A linear transformation layer is used to produce the generated element distribution $p_t$ over the candidate elements $V$:
\begin{equation}
p_t = \text{softmax}(V W_v h_t + b_v),
\end{equation}
where $V$ is composed of the vocabulary and the candidate answers to the user request, $W_v$ and $b_v$ are parameters. 

\subsection{Training}
The training of \ourmodel follows the standard procedure of a sequence-to-sequence model. The BART model is fine-tuned in the training process. Cross-entropy loss is utilized to measure the loss of generating clarification questions and answers.

\section{Data Collection}
\begin{figure}[!t]
\centering
\includegraphics[width=0.48\textwidth]{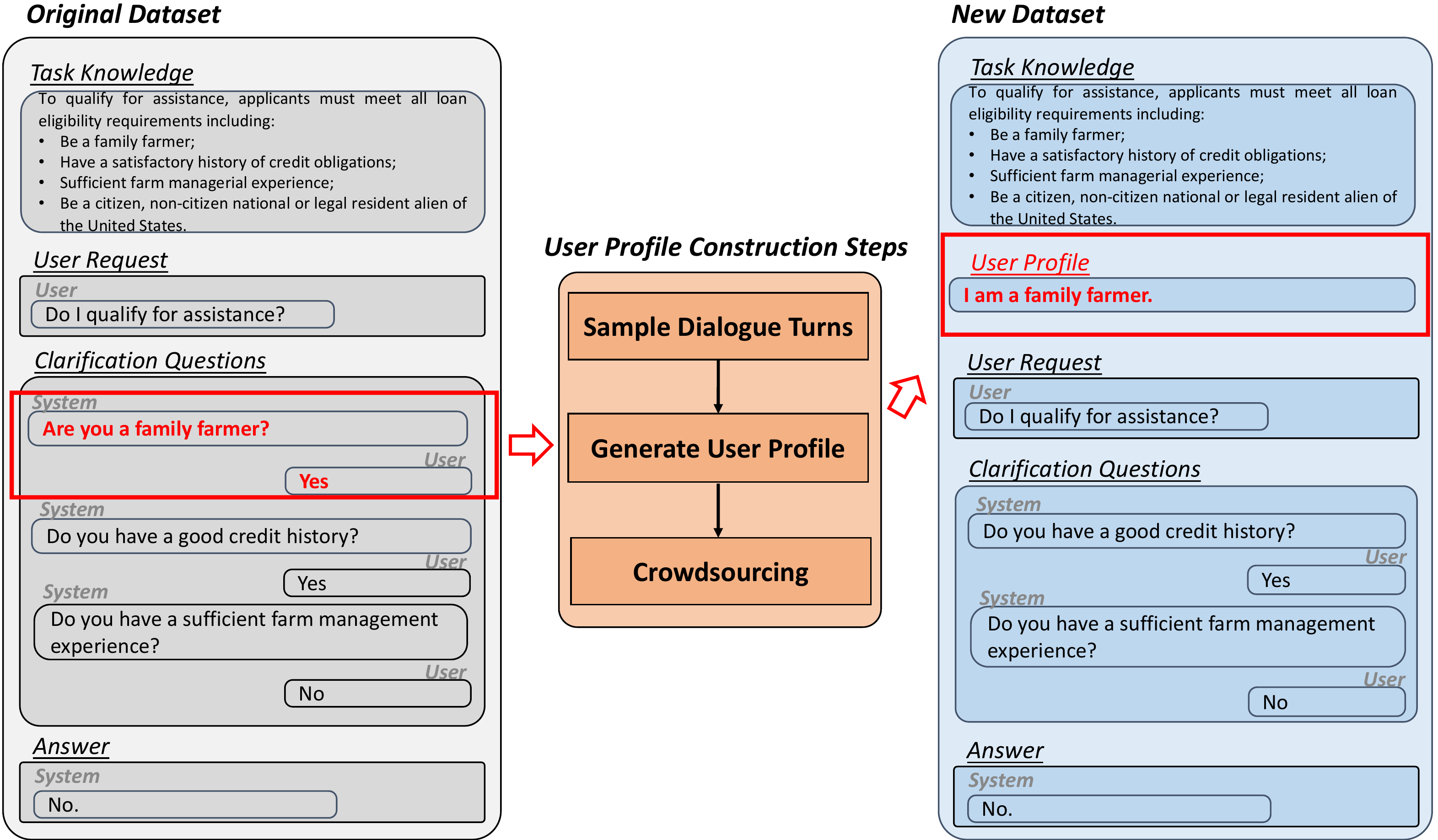}
\caption{The process of user profile construction. We follow a three-step strategy as follows: (1) Sample dialogue turns; (2) Generate user profile; and (3) Correct grammatical errors through crowdsourcing.}
\label{fig:data}
\end{figure}

In task-oriented information seeking, to provide accurate and personalized answers, the system needs to clarify user requests and user profiles based on task knowledge. To the best of our knowledge, no datasets have been developed for this purpose. The closest dataset that can be used is the ShARC\cite{saeidi2018interpretation} dataset, which asks clarification questions based on task knowledge to clarify user information-seeking requests. The left example in Figure~\ref{fig:data} is from the ShARC dataset. However, the ShARC dataset does not contain user profiles, which can not be directly used in our settings. To build a suitable dataset for our setting, we need to extend each dialogue in ShARC dataset with a user profile. The new extended dataset is called \ourdataset. This is the first public dataset that focuses on asking clarification questions for information seeking on task-oriented dialogues. In this section, we explain how we extend the ShARC dataset for task-oriented information seeking.

\subsection{User Profile Construction}
\label{user_profile}
Because the original ShARC dataset lacks personalized information, we need to construct a user profile for each dialogue. Many researchers have demonstrated that dialogue context usually contains strong suggestions for personalized information~\citep{zhong2022less, wu2020getting}. Therefore, following previous work~\citep{pei2021cooperative, qian2021learning}, we also utilize dialogue context to construct user profiles. As shown in Figure~\ref{fig:data}, we use a three-step strategy to construct user profiles. In the first step, we sample some dialogue context. In the second step, we rewrite the sampled dialogue context into declarative sentences as the user profile using rules. In the third step, we use human annotators to correct any grammatical errors incurred during the rewriting process.


Specifically, we first randomly sample 1-5 dialogue turns from the dialogue context. Each dialogue turn in the original dialogue context consists of a clarification question and a corresponding answer. We use the sampled clarification question and answer pairs to construct the user profile. Given the generated user profile will contain the answer to these clarification questions, these clarification questions are unnecessary to be asked in the dialogue. Therefore, we remove these clarification questions from the original dialogue context to improve the efficiency of the dialogue.

After we sample dialogue turns, we need to rewrite them into declarative sentences to generate user profiles. For each dialogue turn, we identify the auxiliary verb in the clarification questions (such as ``Are'', ``Do'', etc.) and the polarity of the answers (such as ``Yes'' and ``No''). Based on the auxiliary verb and polarity, we use rules to map each clarification question and answer pair into a declarative sentence. For example, in Figure~\ref{fig:data}, the clarification question and answer pair is ``Are you a family farmer? Yes''. The rewritten sentence is ``I am a family farmer.''
Finally, the user profile is constructed by concatenating all these rewritten declarative sentences.

However, this rewriting process is not always as straightforward; in these cases, to improve the quality of the generated user profiles, we collect human annotations to correct grammatical errors incurred during the rewriting process. It is important to note that our annotators not only check grammatical errors in the rewriting declarative sentences but also are required to provide suitable corrections for the errors as well. 
All the grammatical errors in the rewriting of declarative sentences are replaced with the corrections provided by the annotators.
After that, we have a new dataset \ourdataset that is suitable for task-oriented information-seeking problems. We split \ourdataset into train, development, and test sets such that the train set includes 70\% of the conversations, the development set contains 10\% of them, and the rest 20\% is the test set. Details about the \ourdataset dataset are shown in Table \ref{tbl:dataset}.

\begin{table}[!t]
\centering
\caption{Number of dialogues, task knowledge, user profiles, and turns in the training, validation and testing sets of \ourdataset.}
\resizebox{0.49\textwidth}{!}{
\begin{tabular}{l|cccc}
\toprule
\textbf{Set} & \textbf{\#Dialogue} & \textbf{\#Task Knowledge} & \textbf{\#User Profile} & \textbf{\#Turns}  \\
\midrule
\textbf{All}   & 108,599      & 1,742   & 85,749 & 260,924  \\ 
\textbf{Training} & 76,019       & 687     & 55,048 & 184,027  \\
\textbf{Validation}   & 10,860       & 495     & 10,545 & 25,473    \\
\textbf{Testing}  & 21,720       & 560     & 20,156 & 51,424   \\
\bottomrule
\end{tabular}
}
\label{tbl:dataset}
\end{table}

\subsection{Dataset Quality Check}
We conduct a further human evaluation to assess the quality of \ourdataset. Following previous work~\citep{eric2019multiwoz}, three annotators were asked to evaluate the quality of \ourdataset. 
The criterion for dataset quality evaluation contains five dimensions:
1) Fluency: Is the user profile grammatically well-formed? 
2) Usefulness: Does the user profile have useful personalized information?
3) Relevancy: Is the dialogue context relevant to user request and user profile? 
4) Clarification: Does the dialogue context clarify unclear information in the user request and user profile?
5) Naturalness: Since we removed the sampled dialogue turns from the dialogue context, the naturalness of the dialogues in \ourdataset may be worse than the original dialogues. 
Therefore, we conducted a comparative study where we show annotators one dialogue from \ourdataset and one original dialogue, by asking annotators to identify which dialogue is more natural. 


We randomly sampled 100 dialogues from \ourdataset. 
Under fluency, usefulness, relevancy, and clarification dimensions, the ratios of dialogues that are satisfied with the corresponding dimension are all 1.0.
For the naturalness dimension, the ratio of identifying \ourdataset is the more natural dialogues is approximately equal to 0.5. This indicates that \ourdataset is as natural as the original ShARC dataset. The evaluation results on relevancy, usefulness, fluency, clarification, and naturalness indicate the high quality of \ourdataset.


\section{Experiments}



\subsection{Baselines}
We compare our approach with the following state-of-the-art baselines,
which include clarification question generation methods and task-oriented dialogue system response generation methods. Since the baselines cannot utilize task knowledge and user profile information, to make the comparison between \ourmodel and baselines fair, we concatenate the task knowledge, user profile, and dialogue context together as the inputs of the baselines.

\begin{itemize}
\item \partitle{GAN-Utility}~\cite{rao2019answer}: State-of-the-art on clarification question generation, which is a sequence-to-sequence generative network for generating clarification questions in open-domain dialogues.

\item \partitle{SOLOIST}~\cite{peng2021soloist}: State-of-the-art on task-oriented dialogue system response generation, which uses a transformer-based auto-regressive language model to generate system responses. Since we don't have the dialogue states and dialogue actions in \ourdataset, this model is only trained on the loss of dialogue system response generation.

\item \partitle{UBAR}~\cite{yang2021ubar}: State-of-the-art on task-oriented dialogue system response generation, which utilizes the large pre-trained unidirectional language model GPT-2 to generate system responses on the sequence of the entire task-oriented dialogue session. Similar to SOLOIST, this model is only trained on the loss of dialogue system response generation.
\end{itemize}

\subsection{Evaluation Measures}

\partitle{Clarification Question Generation.} Following previous work on clarification question generation~\cite{rao2019answer,peng2021soloist,yang2021ubar,majumder2021ask}, we use both automatic metrics and human evaluation to evaluate our approach.

The automatic metrics we used are as follows:
\begin{itemize}
\item \partitle{BLEU}~\cite{papineni-etal-2002-bleu} estimates a generated clarification question via measuring its n-gram precision against the ground truth. 
\item \partitle{ROUGE}~\cite{lin-2004-rouge} measures n-gram recall between generated clarification question and ground truth. 
\end{itemize}

The human evaluation considers the following dimensions: 
\begin{itemize}
\item \partitle{Fluency}: Is the clarification question grammatically well-formed? 
\item \partitle{Relevance}: Is the clarification question relevant to the user request and user profile? 
\item \partitle{Clarification}: Does the clarification question clarify unclear information in the user request and user profile? 
\item \partitle{Usefulness}: We also perform a comparative study where we show annotators two clarification questions along with the dialogue by asking the annotators to choose which of the clarification questions is more useful to solve the user request.
\end{itemize}


\partitle{Answer Prediction.}
Similar to previous work on information seeking~\citep{rao2019answer,majumder2021ask}, we use \textit{Success} to evaluate our approach. 
\begin{itemize}
\item \partitle{Success} is calculated by measuring how often the dialogue system provides the right answer to the user request.  
\end{itemize}


\begin{figure*}
  \centering
  \subfloat[Success]
    {
    \includegraphics[scale=0.33]{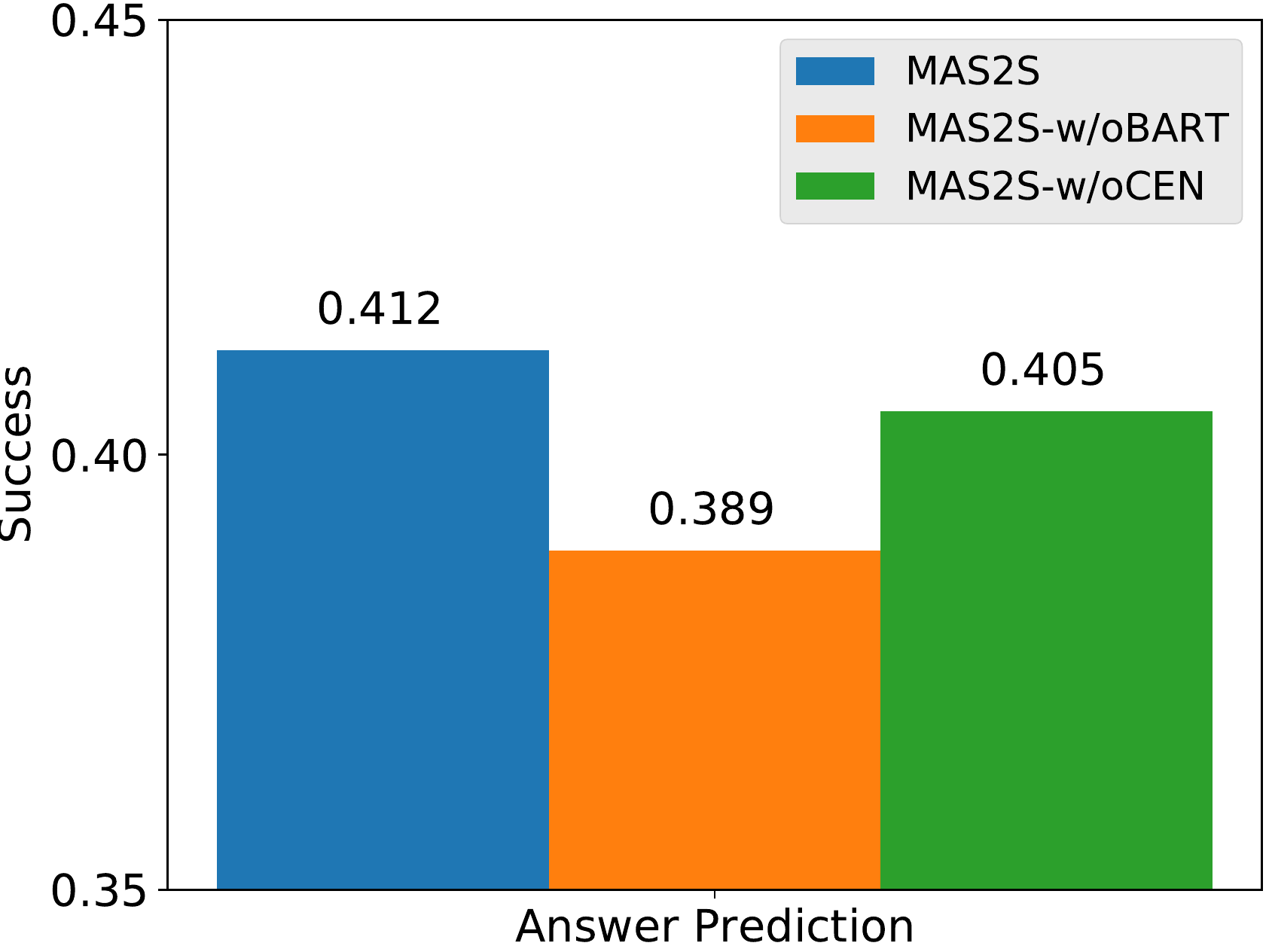}
    \label{fig:ablation_acc}
    }
  \subfloat[BLUE]
    {
    \includegraphics[scale=0.33]{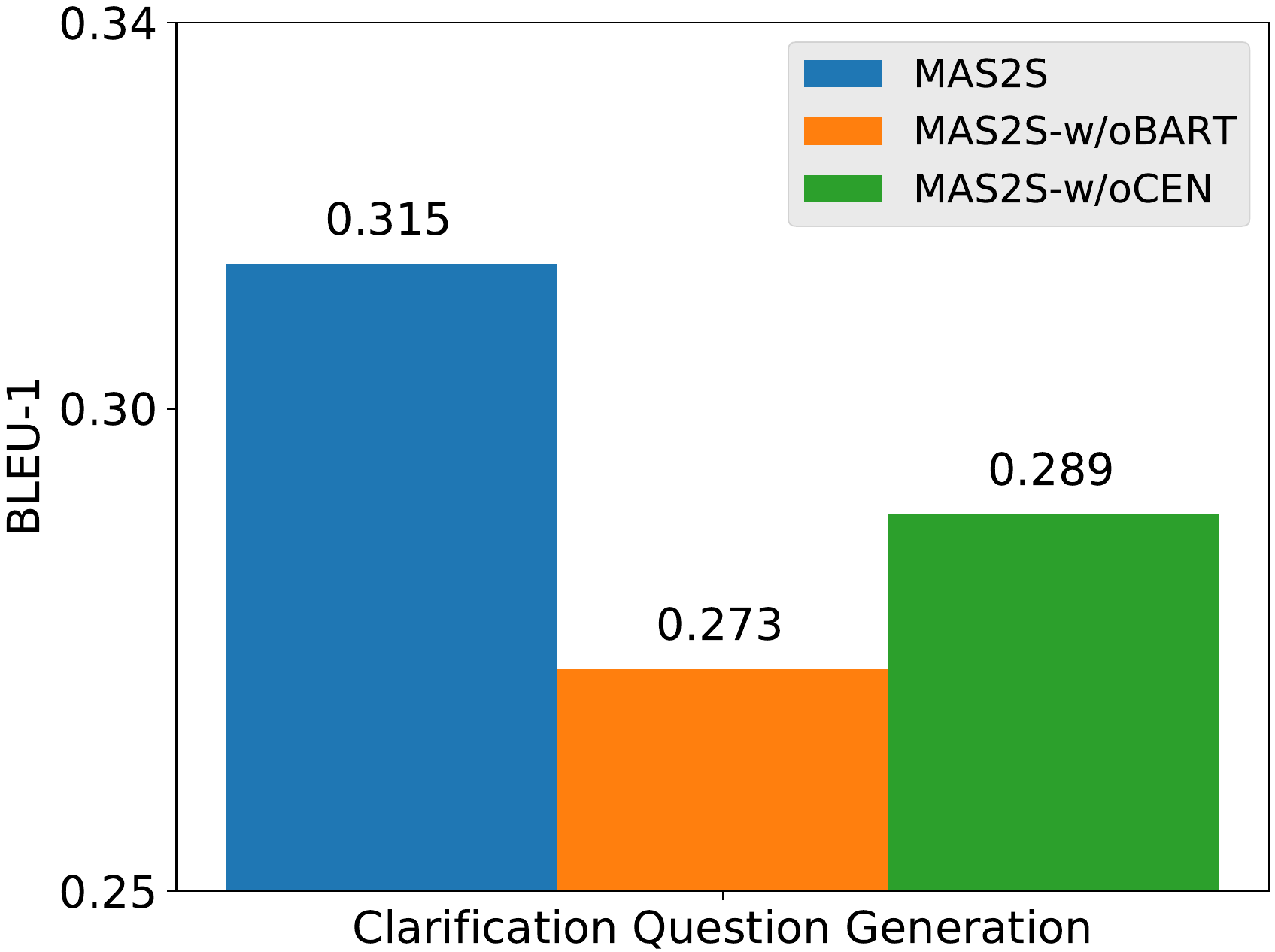}
    \label{fig:ablation_bleu}
    }
  \subfloat[ROUGE]
    {
    \includegraphics[scale=0.33]{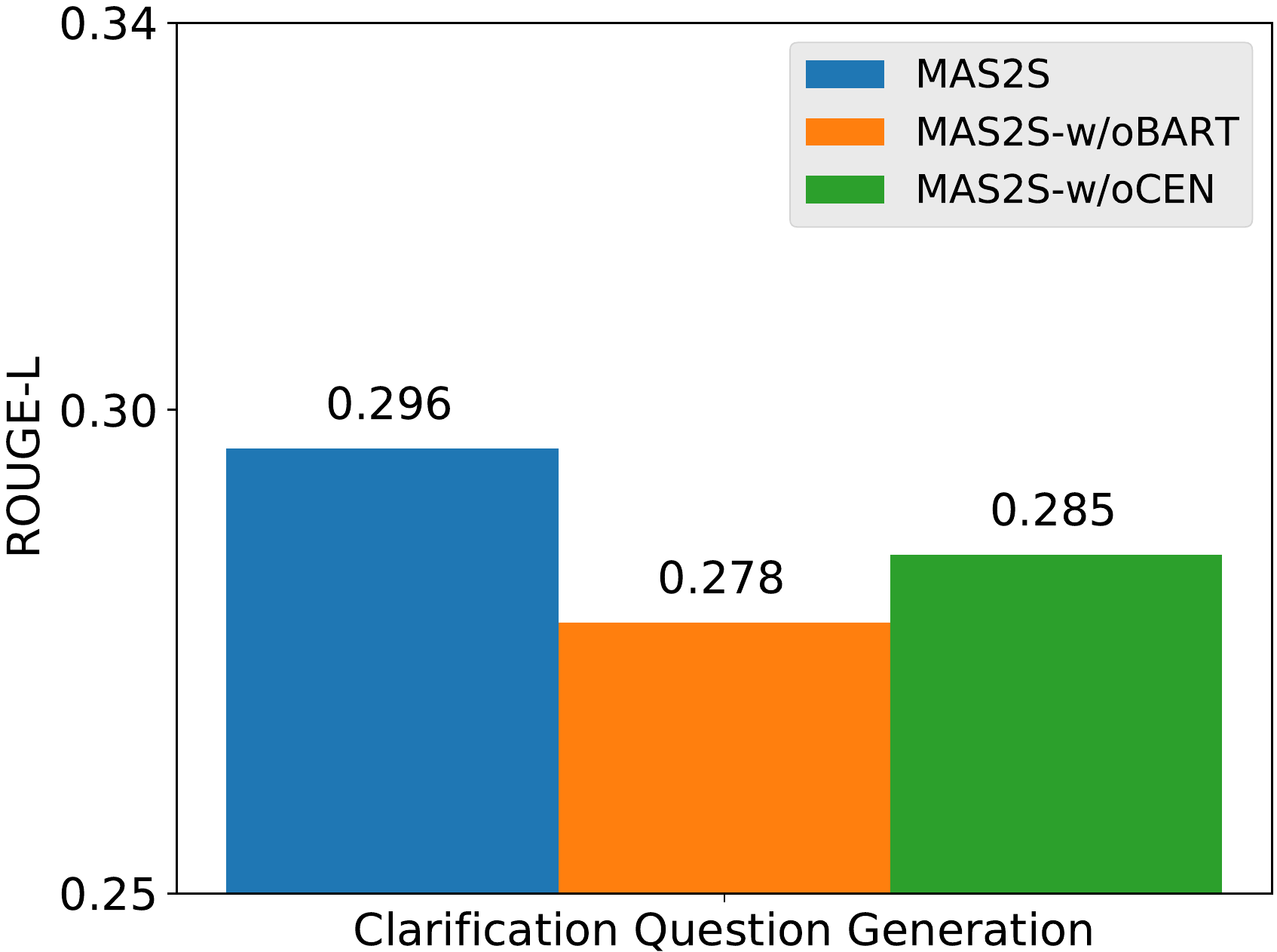}
    \label{fig:ablation_rough}
    }
\caption{Ablation study of \ourmodel with respect to BART, and confidence embeddings network on \ourdataset.}
\label{fig:ablation}
\end{figure*}

\begin{figure*}
  \centering
  \subfloat[Example 1]
    {
    \includegraphics[scale=0.37]{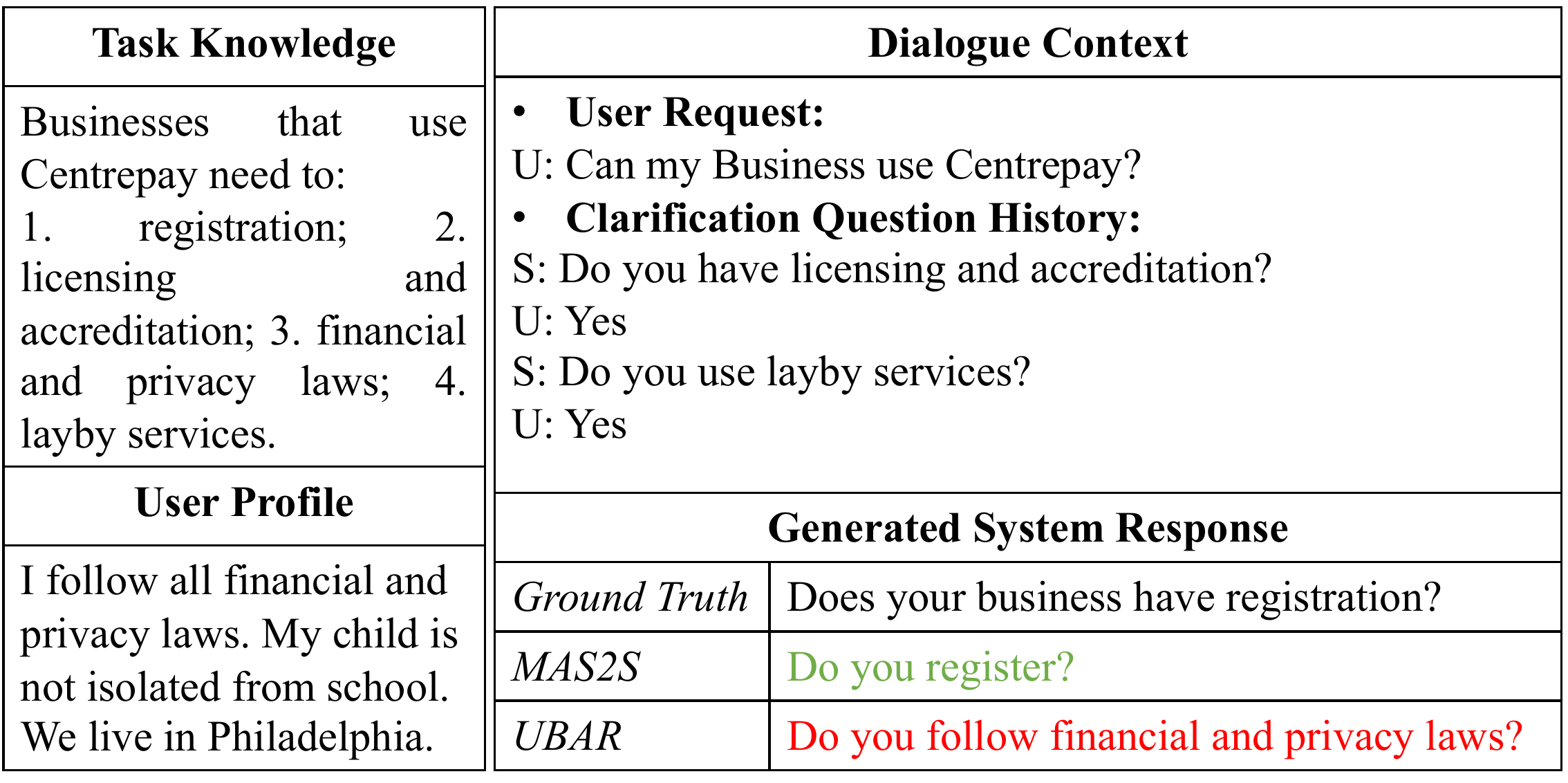}
    \label{fig:ablation_bleu}
    }
  \subfloat[Example 2]
    {
    \includegraphics[scale=0.37]{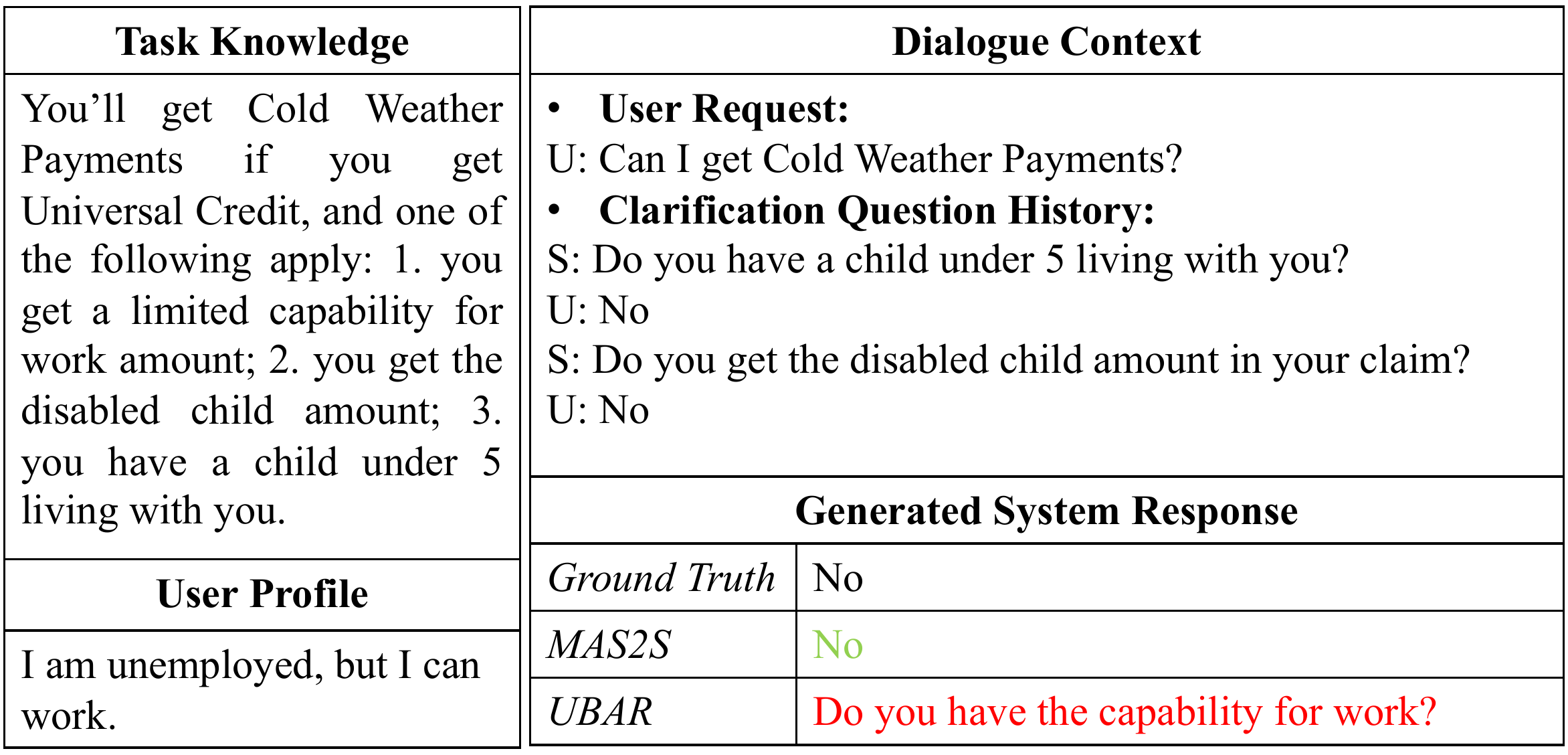}
    \label{fig:ablation_rough}
    }
\caption{Case study based on \ourmodel and the best baseline UBAR on \ourdataset. The generated system response in green is correctly predicted, while the generated system response in red is incorrectly predicted.}
\label{fig:case}
\end{figure*}


\subsection{Implementation Details}
We use a pre-trained BART-based model to encode dialogue context, user profiles, and task knowledge. The BART model is fine-tuned in the training process. The max sentence length of BART is set to 512. The hidden size of all the attention layers is set to 768. We also use beam search for decoding, with a beam size of 5. The dropout probability is set to 0.1. The batch size is set to 4. We optimize with Adam optimizer ~\cite{kingma2014adam} and an initial learning rate of 1e-4. Hyperparameters are chosen using the validation dataset in all cases.

\subsection{Experimental Results}
For the clarification question generation task, the automatic metric evaluation results and the human evaluation results on \ourdataset are shown in Table~\ref{tab:results1} and~\ref{tab:results2}. From the results, we can see that \ourmodel performs significantly better than the baselines on clarification question generation in terms of BLEU, ROUGE, and human evaluation criterion. The results indicate that \ourmodel can effectively leverage the task knowledge to clarify the user requests and the user profiles for task-oriented information seeking. 

For the answer prediction task, the experiment results on \ourdataset are shown in Table~\ref{tab:results1}. We can observe that \ourmodel performs significantly better than the baselines in terms of Success. The results indicate that \ourmodel, which contains the task knowledge attended answer confidence embedding network, is able to provide much more accurate answers for task-oriented information seeking.

\begin{table}[!t]
\centering
\caption{Performance of \ourmodel and baselines on automatic metrics. Numbers in \textbf{bold} denote best results in that metric. \ourmodel significantly improves over the best baseline (two-sided paired t-test, p < 0.05).}
\resizebox{0.35\textwidth}{!}{
\begin{tabular}{l|ccc}
        \toprule
        \textbf{Model}&\textbf{Success}&\textbf{BLEU}&\textbf{ROUGE} \\
 		\midrule
        \text{GAN-Utility}& 0.334 &0.224 &0.216  \\
        \text{SOLOIST} &0.352&0.229  &0.219  \\
        \text{UBAR} &0.397& 0.282  &0.273  \\
        \midrule
        \text{\bf \ourmodel} & \textbf{0.412}& \textbf{0.315} & \textbf{0.296}\\
		\bottomrule
	\end{tabular}
	}
\label{tab:results1}
\end{table}

\begin{table}[!t]
\centering
\caption{Human evaluation results of \ourmodel and the best baseline UBAR on 100 randomly sampled dialogues from \ourdataset dataset. Numbers in \textbf{bold} denote the best results in that metric. MAS2S significantly improves over the best baseline UBAR (two-sided paired t-test, p < 0.05).}
\resizebox{0.49\textwidth}{!}{%
\begin{tabular}{l|cccc}
        \toprule
        \textbf{Model}&\textbf{Fluency}&\textbf{Relevance}&\textbf{Clarification}&\textbf{Usefulness}\\
        \midrule
        \text{UBAR} & 0.59&  0.72&  0.48&  0.29\\
        \text{\bf \ourmodel} & \textbf{0.61}& \textbf{0.78} & \textbf{0.58} & \textbf{0.71}\\
		\bottomrule
	\end{tabular}
	}
\label{tab:results2}
\end{table}



\section{Discussions}
\label{sec:results}

\subsection{Ablation Study}
We also conduct an ablation study on \ourmodel. We validate the effects of two factors: confidence embeddings network and BART-based encoder/decoder. The results indicate that all the components of \ourmodel are indispensable.

\partitle{Effect of Confidence Embeddings Network.}
To investigate the effectiveness of using the confidence embeddings network, we compare \ourmodel with \ourmodel-w/oCEN, which eliminates the confidence embeddings network module. We concatenate the semantic embeddings of dialogue context, user profile, and task knowledge as the initial state of the decoder. Figure~\ref{fig:ablation} shows the results on \ourdataset in terms of BLEU-1, ROUGE-L, and Success. From the results, we can see that the performances of both clarification question generation and answer prediction deteriorate considerably without the confidence embeddings network. This indicates that the confidence embeddings network helps provide a more accurate indication about when and how to ask clarification questions and predict answers.

\partitle{Effect of BART.}
To investigate the effectiveness of using BART in the text encoder, and natural language decoder, we replace BART with the standard sequence-to-sequence Transformer~\citep{vaswani2017attention} and run the model on \ourdataset. As shown in Figure~\ref{fig:ablation}, the performances on both clarification question generation and answer prediction of the Transformer-based model \ourmodel-w/oBART decreases significantly compared with \ourmodel, in terms of BLEU-1, ROUGE-L, and Success. It indicates that the BART-based encoder/decoder can create and utilize more accurate representations for dialogue, user profile, and task knowledge on both clarification question generation and answer prediction.



\subsection{Case Study}
We qualitatively analyze the results of \ourmodel and the best baseline UBAR on \ourdataset dataset. We find that \ourmodel generates more accurate system responses by leveraging the relation existing in the dialogue context, user profile and task knowledge. For example, in the first case in Figure~\ref{fig:case}, the user profile mentions that \textit{``I follow all financial and privacy laws''}. \ourmodel can correctly infer the system needs to ask another clarification question about \textit{``registration''} instead of \textit{``financial and privacy laws''} considering the task knowledge and dialogue context. In the second case, the user profile mentions that \textit{``I can work''}. \ourmodel can correctly provide the answer \textit{``No''} instead of asking a clarification question about \textit{``capability for work''} also considering the task knowledge and dialogue context. From the examples above, we can see that \ourmodel can effectively extract the relation between dialogue context, user profile, and task knowledge, yielding correct system responses. In contrast, UBAR can not model these relations. Thus it cannot properly generate system responses.


\subsection{Utility of Clarification Questions}
\label{sec:appendix_a}
\begin{figure}[!th]
\centering
\includegraphics[width=0.45\textwidth]{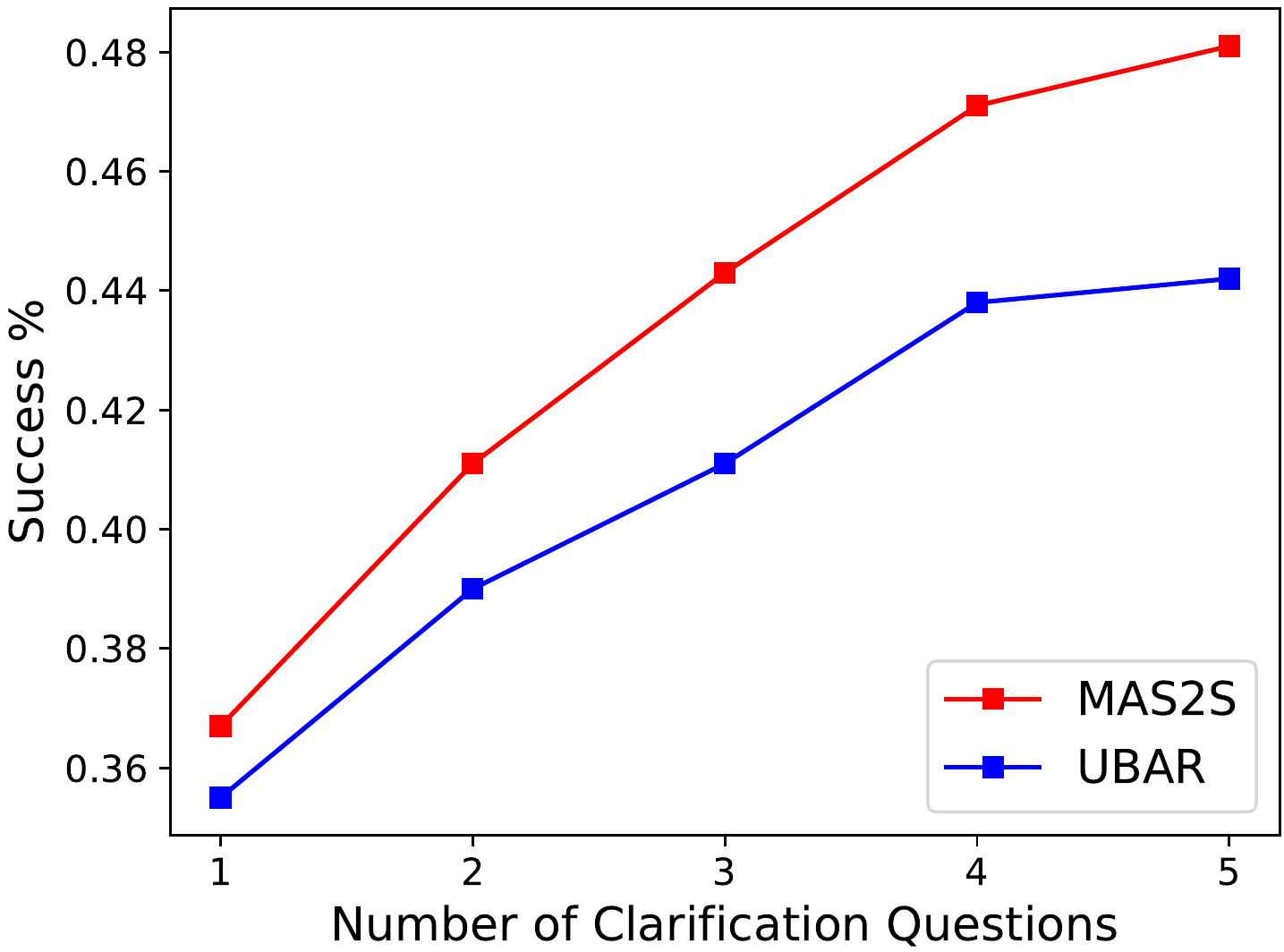}
\caption{Impact of number of clarification questions on the performance of Success of \ourmodel and the best baseline UBAR.}
\label{fig:turn}
\end{figure}

To investigate the utility of the generated clarification questions, we compare the performance of Success of \ourmodel with the best baseline UBAR on dialogues asking different number of clarification questions. Figure~\ref{fig:turn} shows the performance of Success of \ourmodel and UBAR on dialogues with $k$ clarification questions ($k \in \{1,2,3,4,5\}$). The results show that \ourmodel performs better in answer prediction on all dialogues. It indicates the usefulness of the clarification questions generated by \ourmodel. In addition, we can also see that the performance of \ourmodel improves as the dialogue advances to multiple clarification questions. It indicates that appropriate clarification questions can effectively clarify the user request to provide more accurate answer. We conjecture that the task-oriented dialogue system can understand the users information needs and the user profile more accurate with the increase in the number of clarification questions. 

\subsection{Necessary of Clarification Questions}
\label{sec:appendix_b}
We conduct experiments to investigate whether the generated clarification questions are necessary. We compare \ourmodel, the best baseline UBAR in terms of the average number of clarification questions (NoQ) over the dialogues, 
and the corresponding average absolute differences with the number of clarification questions needed (AbsDiff).
As shown in Table~\ref{tab:necessary}, the average number of clarification questions (NoQ) of \ourmodel is less than the best baseline UBAR. And when comparing average absolute differences with the number of clarification questions needed (AbsDiff), we observe that the number of clarification questions of \ourmodel is more closer to the ground truth number of clarification questions (Oracle) than the best baseline UBAR. It indicates that \ourmodel can generate more necessary questions to clarify the user request and user profile than the best baseline UBAR. 

\begin{table}[!t]
\centering
\caption{Average number of clarification questions (NoQ) and absolute difference of clarification questions (AbsDiff) over the dialogues of \ourmodel and the best baseline UBAR on the test set of \ourdataset. Numbers in \textbf{bold} denote best results in that metric.}
\resizebox{0.25\textwidth}{!}{
\begin{tabular}{l|cc}
\toprule
\textbf{Model}&\textbf{NoQ} &
\textbf{AbsDiff}\\
\midrule
\text{Oracle} & 2.36 & 0.00 \\
\text{UBAR} & 3.14 & 1.68 \\
\text{\bf{\ourmodel}} & \bf{2.97}  & \bf{0.94} \\
\bottomrule
\end{tabular}
}
\label{tab:necessary}
\end{table}

\subsection{Impact of User Request Length}
\label{sec:appendix_c}
\begin{figure}[!th]
\centering
\includegraphics[width=0.45\textwidth]{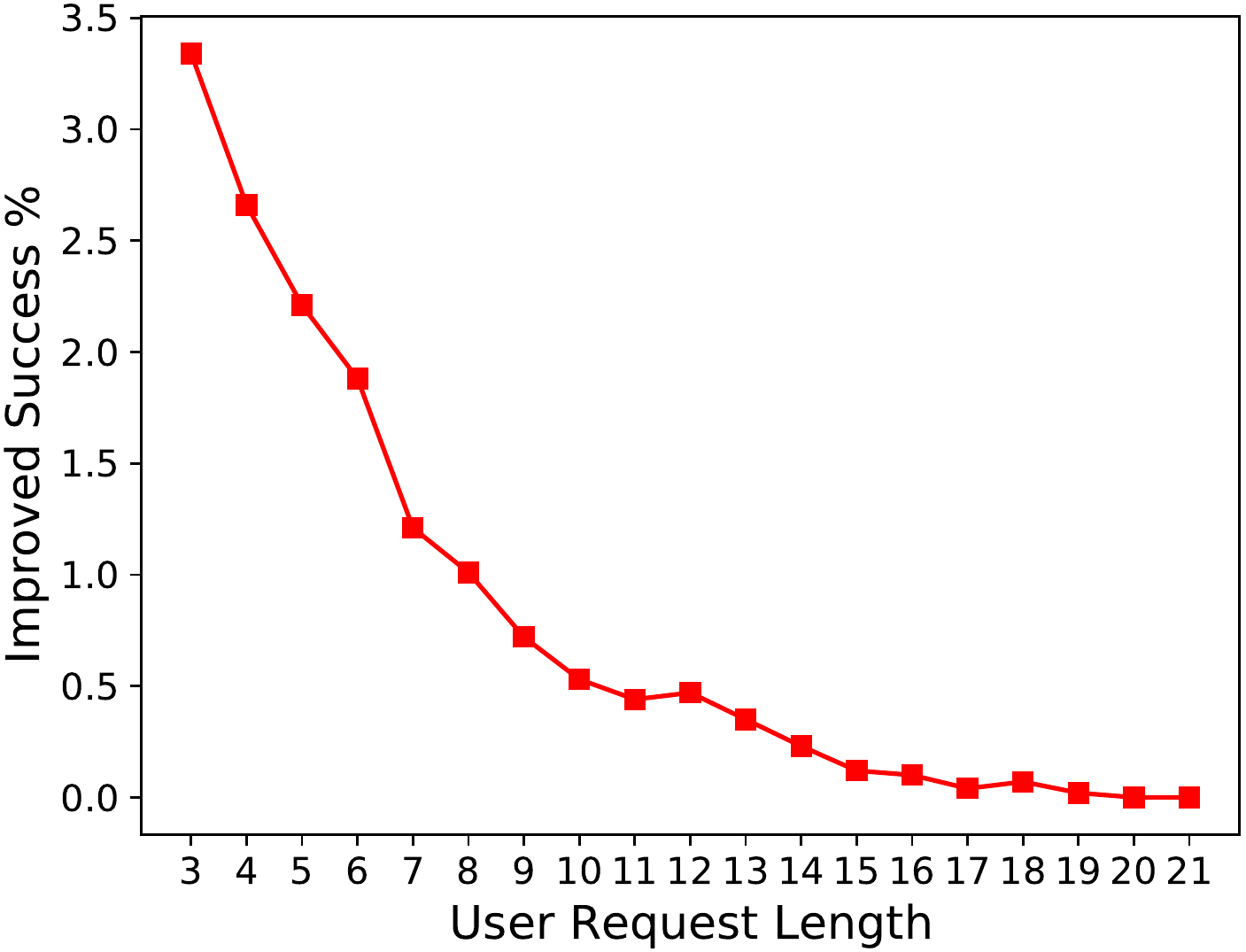}
\caption{Impact of user request length on the performance of Success of \ourmodel.}
\label{fig:length}
\end{figure}

We analyze the performance of \ourmodel based on the number of user request tokens. Figure~\ref{fig:length} shows the improved Success of \ourmodel on different lengths of user request compared to the best baseline UBAR. From the results, one can observe that \ourmodel performs better in all cases, no matter the length of the user request. It indicates that utilizing the clarification question generation model in task-oriented dialogue is necessary to clarify user request. In addition, the relative improvement of \ourmodel is negatively correlated with the length of the user request. It indicates that the shorter user request needs clarification in more cases. We conjecture that it is due to the shorter user request usually containing more ambiguous information. Asking clarification questions can effectively improve the natural language understanding ability of shorter user request in task-oriented dialogues.

\subsection{Impact of User Profile}
\label{sec:appendix_d}
To investigate the effectiveness of user profiles, we compare \ourmodel with \ourmodel-w/oProfile, which eliminates the user profile and knowledge attended user profile representations. Table~\ref{tab:user} shows the results on \ourdataset in terms of BLEU-1, ROUGE-L, and Success. From the results, we can see that without user profiles, the performance of the clarification question generation deteriorates considerably. It indicates that the relation between the user profiles and the user requests helps the dialogue systems understand the user information needs. In addition, the performance of Success also deteriorates without using user profiles. We conjecture that it is due to the dialogue system's lack of personalization information to provide an accurate response to users. Thus, the utilization of user profiles is desirable.

\begin{table}[!t]
\centering
\caption{Performance of \ourmodel and \ourmodel-w/oProfile on \ourdataset dataset. Numbers in \textbf{bold} denote best results in that metric. MAS2S significantly improves over the \ourmodel-w/oProfile (two-sided paired t-test, p < 0.05).}
\resizebox{0.49\textwidth}{!}{
\begin{tabular}{l|c|c|c}
        \toprule
        \textbf{Model}&\textbf{Success}&\textbf{BLEU}&\textbf{ROUGE}\\
        \midrule
        \text{\ourmodel-w/oProfile} & 0.381& 0.256 & 0.264 \\
        \text{\bf \ourmodel} &\textbf{0.412}& \textbf{0.315} & \textbf{0.296} \\
		\bottomrule
	\end{tabular}
	}
\label{tab:user}
\end{table}

\section{Conclusion and Future Work}
\label{sec:conclusion}
In this work, we focused on the problem of task-oriented information seeking. We proposed a Multi-Attention Seq2Seq Networks (\ourmodel) to generate clarification questions and predict answers for task-oriented information seeking, which integrates the power of both sequential modeling and attention mechanisms. Due to no existing dataset suitable for task-oriented information seeking, we also constructed and released a new dataset called \ourdataset, which includes accurate and personalized clarification questions for task-oriented information seeking. 
Experiments on \ourdataset verified the performance of \ourmodel against state-of-the-art clarification question generation baselines and task-oriented dialogue system response generation baselines. 
The research on asking clarification questions for information seeking on task-oriented dialogues is still in its initial stage, and this work is just one of the first steps. In the future, the proposed paradigm may also be extended to more complex scenarios, such as considering task relations, dialogue relations, multi-modal information, etc.


\bibliographystyle{ACM-Reference-Format}
\bibliography{sample-base}

\appendix

\end{document}